\documentclass[letterpaper, 10pt, journal, twoside]{IEEEtran}

\usepackage{iftex}
\ifXeTeX
  \usepackage[T1]{fontenc}
\fi
\usepackage{amsmath,amssymb}
\usepackage{graphicx}
\usepackage{booktabs}
\usepackage{xcolor}
\usepackage{cite}
\usepackage{siunitx}
\usepackage[hidelinks]{hyperref}

\newcommand{\projecturl}{\url{https://woojin8098.github.io/CoTAMP/}}

\begin{document}

\title{CoTAMP: Cooperative Task and Motion Planning for Quadruped Robots with Large Language Models}

\author{Woojin Kim$^{1}$ and Jemin Hwangbo$^{1,*}$%
\thanks{$^{1}$The authors are with the Department of Mechanical Engineering, Korea Advanced Institute of Science and Technology (KAIST), Daejeon 34141, Republic of Korea. \texttt{\{woojin0624, jhwangbo\}@kaist.ac.kr}}%
\thanks{$^{*}$Corresponding author.}}

\maketitle

\begin{abstract}
Some environments cannot be traversed by a single quadruped robot, for example a ledge that one robot can climb only from its partner's back or a gap that must be bridged with boxes. Solving them requires cooperative task and motion planning (TAMP). Previous works are limited to the motions of a single robot or only select actions from a predefined set of skills. We propose a framework with two parts: task planning that invents motions not defined in advance, and motion generation that realizes the invented motions. Both parts consist of large language model (LLM)-based modules and code-based modules. Task planning divides the trajectory of the whole task into steps and fixes the robot poses at the step boundaries. Motion generation produces keyframes that describe the motion of each step and turns them into a continuous trajectory through kinematic synthesis. The code-based modules check every plan and motion and return failures to the LLM-based modules for correction. The whole trajectory is then refined by trajectory optimization to satisfy the dynamics. On three simulated courses with two RAIBO2 robot models, our framework produced kinematically valid references in all 30 trials, all of which converged in trajectory optimization. Compared with several baselines, our framework showed advantages in success rate, dynamic feasibility, execution time, and generation time and cost.

Project page: \projecturl
\end{abstract}


\section{Introduction}
\label{sec:intro}

\begin{figure}[t]
\centering
\includegraphics[width=\linewidth]{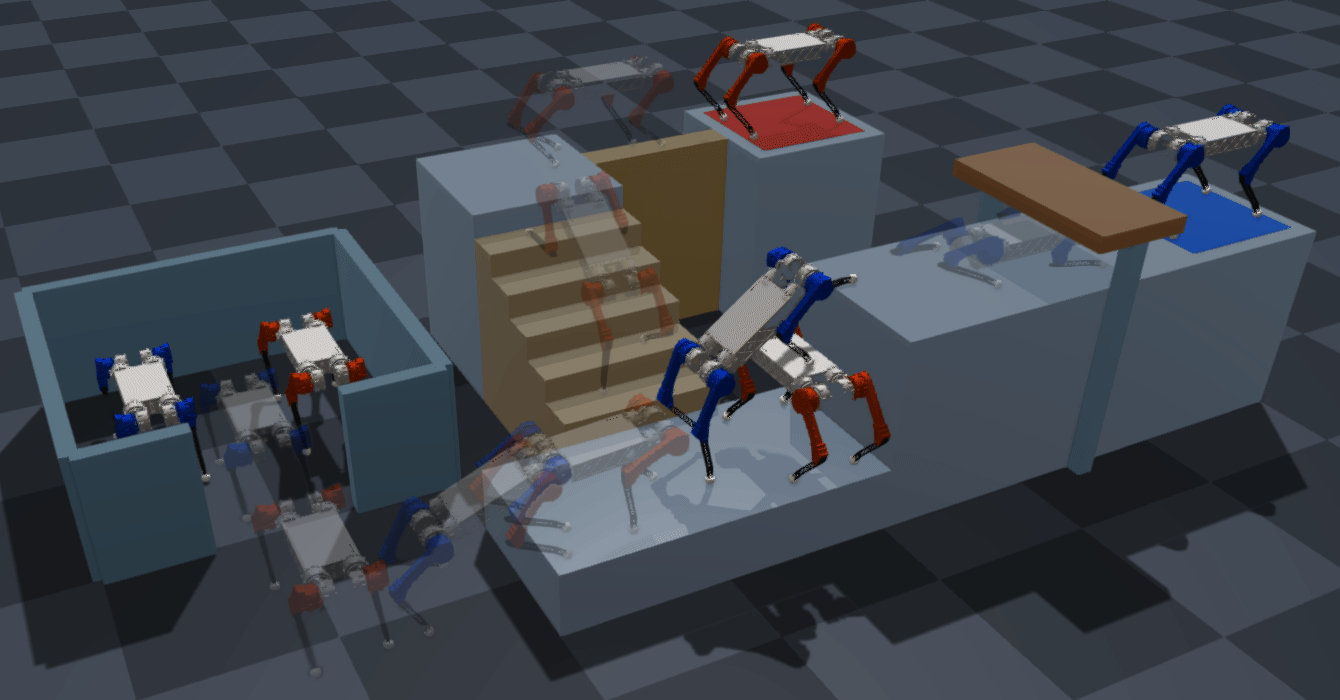}
\caption{A trajectory generated by our framework on Course~1: the blue robot climbs onto the upper platform from the red robot's back, and both robots then reach their goals.}
\label{fig:teaser}
\end{figure}

\IEEEPARstart{T}{ask} and motion planning (TAMP) has enabled legged robots to complete long-horizon tasks in cluttered environments, combining decisions about what to do with motions that respect contacts and geometry~\cite{ouyang2024longhorizon}. Many environments, however, cannot be traversed by a single robot: a ledge higher than its legs can reach, a gap between two platforms, or a path blocked by boxes that must first be pushed into place. Two robots can overcome such obstacles if one serves as a step (Fig.~\ref{fig:teaser}) or a bridge for the other, or if they rearrange objects together. Planning such behavior requires a cooperative TAMP that decides who supports whom and in which order, and at the same time where each foot lands on the partner's back.

\begin{figure*}[t]
\centering
\includegraphics[width=0.32\linewidth]{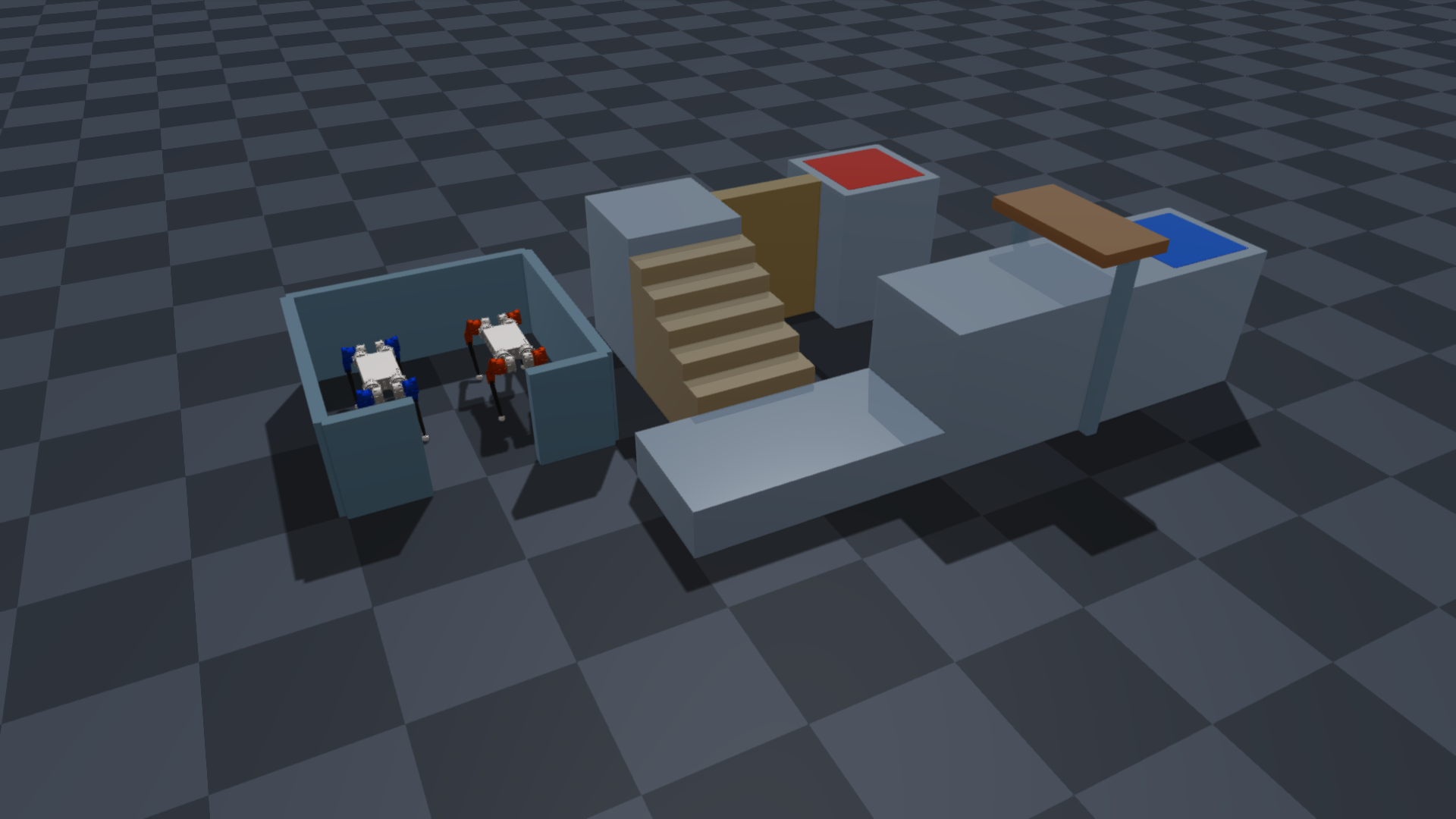}\hfill
\includegraphics[width=0.32\linewidth]{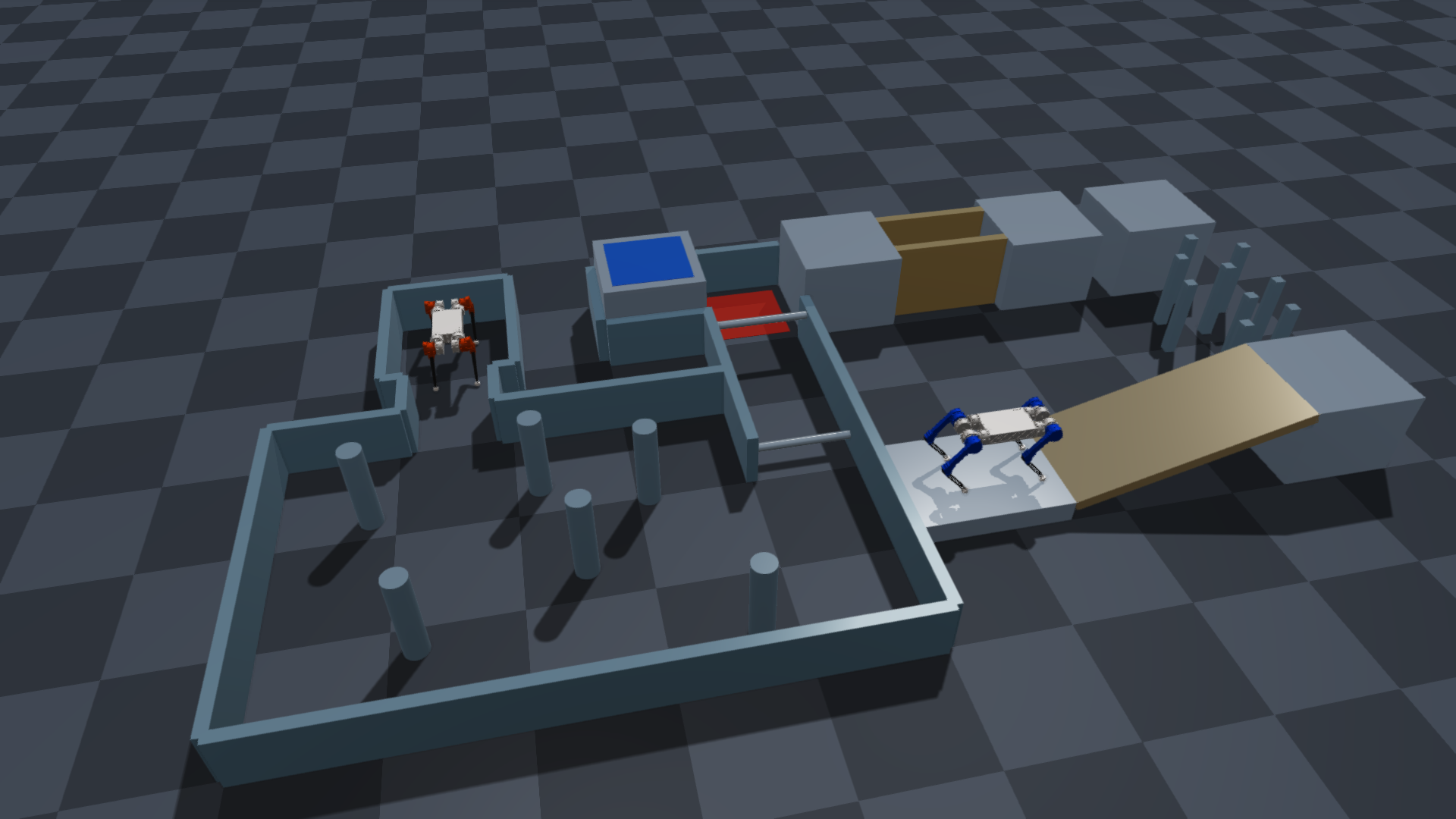}\hfill
\includegraphics[width=0.32\linewidth]{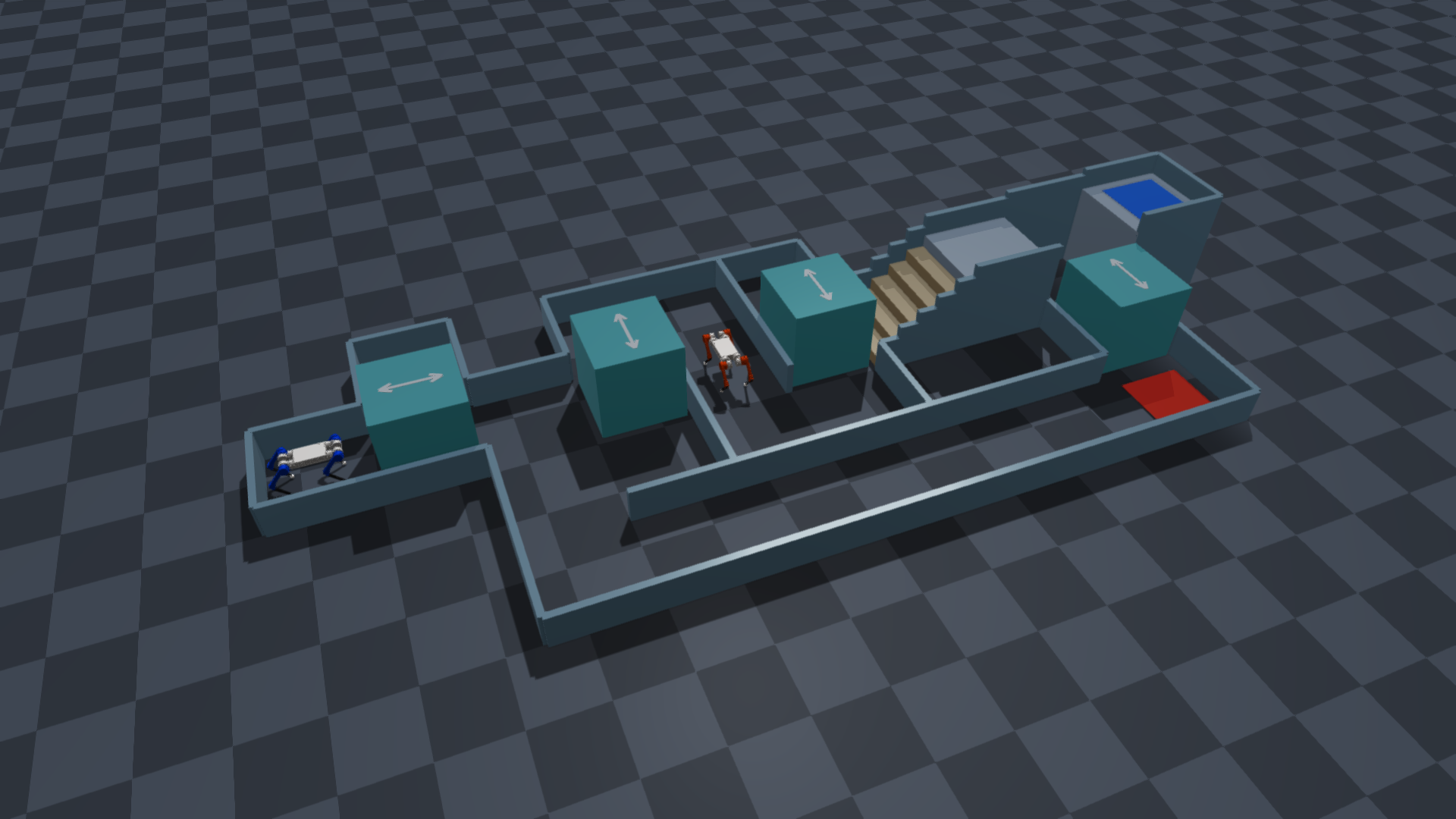}
\caption{The three cooperative courses. Goal regions are colored by the robot that must reach them. (a) Course~1: climbing onto the partner. (b) Course~2: crossing a gap over the partner's back. (c) Course~3: pushing boxes to free a trapped robot and to build a bridge.}
\label{fig:envs}
\end{figure*}

Existing approaches to TAMP fix the set of available actions in advance. Classical TAMP requires hand-designed action schemas with preconditions and effects, and computes the motion of each action with a motion planner~\cite{garrett2021tamp,toussaint2015lgp}. Large language models (LLMs) have made it possible to plan cooperation directly from a description of the scene; Delfaki \emph{et al.}~\cite{delfaki2025legup}, for example, use an LLM to decompose a task so that one quadruped helps another over a high obstacle. Like most LLM-based planners~\cite{ahn2022saycan,lin2023text2motion,mandi2024roco}, however, their method composes the plan from a predefined or pretrained set of skills, in its case \texttt{GoTo}, \texttt{Sit}, and \texttt{Stand} commands. In both cases a task that requires a motion outside the given set cannot be solved, which limits how far such planners generalize to new obstacles.

We therefore let the task planner invent the motions it needs instead of selecting them. The task planner decomposes the task into \emph{steps} according to the kind of motion required and fixes the robot poses at the step boundaries, while the motion within each step is described freely: either a \emph{solo} motion of one robot or a \emph{coupled} motion of both robots in physical contact. Because solo steps of different robots are independent, they can be executed in parallel, which shortens the execution.

Turning such freely invented motions into trajectories is itself difficult. Learning-based controllers need training for every new skill, and trajectory optimization (TO) needs a contact sequence and a good initial guess~\cite{posa2014direct,winkler2018towr}. LLMs can write the reward or cost of such controllers~\cite{yu2023l2r,ma2024eureka}, but these methods target a single robot. Motions in which two robots touch each other, such as stepping onto a partner whose back moves under the load, are particularly hard because the contacts of both robots are coupled.

Recent work shows that LLMs can also generate motions directly, by writing trajectories or programs that produce them~\cite{kwon2024zeroshot,mimicagent2026}. These methods, however, generate the motion of a single robot and do not produce motions in which two robots physically support each other. Our motion generator produces such coupled motions of both robots. To keep generation fast and inexpensive, the LLM specifies only the landed contact states (keyframes), and code-based modules fill in the motion between them with interpolation and a local kinematic optimization, instead of the LLM writing every sample of a long trajectory. The resulting kinematic reference is finally refined by a TO so that it satisfies the dynamics of both robots.

The contributions of this letter are:
\begin{enumerate}
\item An LLM planner that invents solo and coupled motions without a predefined skill set and executes independent motions in parallel.
\item A motion generator that produces coupled motions of two robots in physical contact, in which the LLM outputs only keyframes and code-based modules fill in the motion with kinematic optimization, which takes less time and fewer tokens than generating dense trajectories directly.
\item An evaluation on three cooperative courses showing that the framework successfully performs cooperative TAMP and has advantages over baselines in success rate, dynamic feasibility, execution time, and generation time and cost.
\end{enumerate}

\begin{figure*}[t]
\centering
\includegraphics[width=\linewidth]{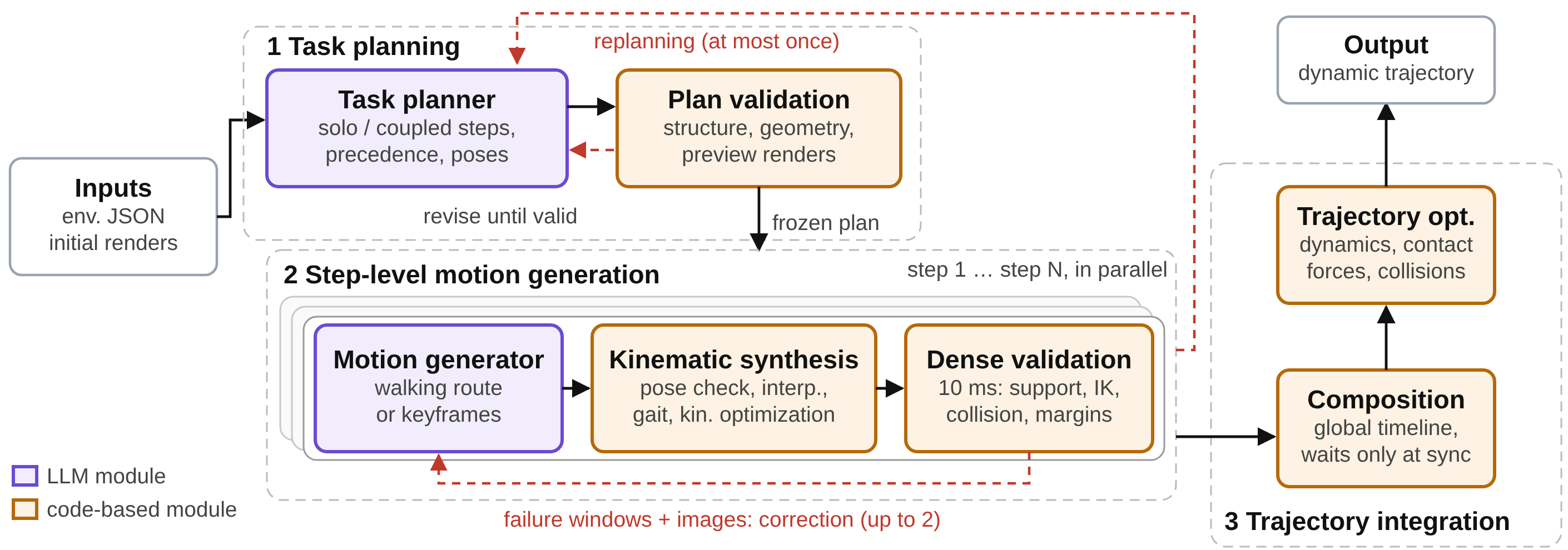}
\caption{Overview of our framework. LLM modules (violet) plan and propose motions; code-based modules (orange) validate, synthesize, and compose them. Steps are generated in parallel, and failures are returned to the LLM for correction or replanning.}
\label{fig:overview}
\end{figure*}

\section{Related Work}
\label{sec:related}

\subsection{LLM-Based Planning}
Classical TAMP combines discrete action search with continuous feasibility reasoning~\cite{garrett2021tamp,toussaint2015lgp}, but assumes hand-designed action schemas. LLM-based planners replace the symbolic search with a language model that selects learned skills~\cite{ahn2022saycan}, writes policy code~\cite{liang2023cap}, checks skill feasibility~\cite{lin2023text2motion}, or coordinates several manipulators~\cite{mandi2024roco}. For quadrupeds, Ouyang \emph{et al.}~\cite{ouyang2024longhorizon} use LLM agents to turn long-horizon locomotion and manipulation tasks into code that calls a set of skills trained with reinforcement learning, and Delfaki \emph{et al.}~\cite{delfaki2025legup} let one robot help another over a high obstacle with \texttt{GoTo}, \texttt{Sit}, and \texttt{Stand} commands. All of these planners select actions from a set defined before planning. Our task planner instead describes each step's motion freely, including motions in which two robots support each other.

\subsection{LLM-Designed Controllers}
A second line of work lets an LLM design the objective of a controller instead of the motion itself. Yu \emph{et al.}~\cite{yu2023l2r} translate instructions into reward parameters for model-predictive control, and Ma \emph{et al.}~\cite{ma2024eureka} write reward functions for reinforcement learning. Wang \emph{et al.}~\cite{wang2024robogen} propose tasks and rewards to learn new skills, and Huang \emph{et al.}~\cite{huang2023voxposer} compose value maps that a motion planner optimizes. Tang \emph{et al.}~\cite{tang2023saytap} map language to foot-contact patterns that a learned locomotion controller executes. These approaches use rewards, value maps, or contact patterns to specify robot behavior, and they consider a single robot. Our focus is on generating mutually consistent contacts and whole-body trajectories for two quadrupeds that physically support one another.

\subsection{Direct Motion Generation}
Other works let a model output the motion directly. Kwon \emph{et al.}~\cite{kwon2024zeroshot} showed that LLMs can generate dense end-effector trajectories for manipulation zero-shot, and Nayak \emph{et al.}~\cite{mimicagent2026} generate quadruped reference trajectories from prompts through planning, code generation, and repair. Other works use generative motion models to produce reference motions from text, human motion, or video that are then tracked by a quadruped controller~\cite{lagoon2024,motionglot2024,unimo2026,kine2go2026}. None of these methods produces motions in which two robots physically interact, such as one robot climbing onto another whose back moves under the load, where the contacts of both robots must be generated consistently and the supporting robot's motion depends on its partner. Our motion generator produces such coupled motions of both robots.

\section{System Overview}
\label{sec:method}

\subsection{Problem Setting}
\label{sec:problem}
We use two RAIBO2 quadrupeds with 12 actuated joints each, distinguished by color (blue and red). An environment consists of static obstacles and movable boxes; each box slides along a rail with one degree of freedom and moves only when a robot pushes it. Each robot must reach the goal region marked in its own color, and the task is achieved when both robots stand inside their goal regions. The three courses used in this paper (Fig.~\ref{fig:envs}) each contain obstacles that a single robot cannot overcome, so every course requires cooperative motions. Given the environment and the initial states, our framework generates whole-body trajectories that bring both robots to their goals while satisfying the dynamics.

\subsection{Pipeline}
The pipeline has a hierarchical structure (Fig.~\ref{fig:overview}). \emph{Task planning} divides the task into steps and fixes their order and the robot poses at the step boundaries. \emph{Motion generation} then produces the motion of each step, with all steps processed in parallel; a final integration step then composes the step motions on a global timeline and refines them with a TO. Task planning and motion generation each combine \emph{LLM modules}, which solve their part by reasoning, and \emph{code-based modules} that we implemented, which check, render, and interpolate the LLM outputs; they are shown in violet and orange in Fig.~\ref{fig:overview}.

\section{Task Planning}
\label{sec:planning}
Task planning consists of the task planner, an LLM module that writes the plan, and code-based modules that validate it.

\subsection{Inputs}
The task planner receives
\begin{itemize}
\item \textit{Environment:} a JSON file with the environment and its obstacles.
\item \textit{Robot models:} the URDF models of both robots.
\item \textit{Images:} renderings of the initial state of the environment.
\item \textit{Output format:} a JSON file that defines the output format.
\item \textit{Prompt:} instructions on what the plan must respect.
\end{itemize}
The full prompts are available on the project page.

\subsection{Plan Representation}
The task planner returns a plan consisting of
\begin{itemize}
\item a step sequence per robot, where a step is either \emph{solo} (the motion of one robot) or \emph{coupled} (the motions of both robots, generated together, with one role each and a stated coupling reason);
\item a free-form behavior family per step;
\item precedence constraints between steps of different robots, i.e., a step may start only after a given step of the other robot has finished;
\item required object states and object coordinate changes per step;
\item the robot positions at the start and end of each step, at which the robots stand with nominal joints and a level base.
\end{itemize}
No skill set is defined in advance: the task planner invents the motions each step needs. The prompt instructs the task planner to consider cooperative motions where an obstacle cannot be overcome by a single robot.

\subsection{Plan Validation}
The code-based modules check the plan in two ways: (1) structural checks verify that the step identifiers and robot assignments are consistent, that the precedence constraints contain no cycles, and that the boxes are moved in a consistent order; and (2) geometric checks place both robots and the boxes in the scene at every step boundary and verify that all four feet of each robot rest on a support, that nothing collides, and that both robots reach their goals at the end.

The failed checks are returned to the task planner, together with renderings of the boundary states that fail a geometric check. The task planner revises the plan based on this feedback, and once the plan passes, it becomes the input of motion generation. This validation covers only the boundary poses; whether the motion between them is feasible is checked during motion generation.

\section{Motion Generation}
\label{sec:motion}
Motion generation consists of the motion generator, an LLM module that writes the motion of each step, and code-based modules that synthesize and validate it. The plan divides the long trajectory of the whole task into short steps. Because the start and end states of every step are fixed by the plan, the steps are independent of each other, so the motions of all steps are generated in parallel, which shortens the generation time.

\subsection{Inputs}
The motion generator of each step receives
\begin{itemize}
\item \textit{Step description:} the start and end states of the step and its description in the plan.
\item \textit{Environment:} the environment JSON file.
\item \textit{Robot models:} the URDF models of the participating robots with a summary of their geometry and joint limits.
\item \textit{Images:} renderings of the start and end states.
\item \textit{Output format:} a JSON file that defines the output format.
\item \textit{Robot interface:} frequently used forward kinematics, inverse kinematics, and collision checks, provided as an API.
\item \textit{Prompt:} instructions for generating the motion of the step.
\item \textit{Motion guidelines:} the constraints every motion must satisfy (support, joint limits, collision avoidance, and edge margins) and the preferred motion style (e.g., continuous body progression and diagonal foot transfers).
\end{itemize}

\subsection{Motion Representations}
The motion generator chooses one of two representations:

\textit{Walking route.} Ordinary walking is comparatively simple and can be generated reliably by rules. For a solo step that only walks or turns on connected horizontal surfaces, the motion generator therefore returns only body waypoints, and a predefined, rule-based walking trajectory generator turns them into a periodic gait at a reference speed of \SI{0.8}{m/s} and \SI{1.4}{rad/s}.

\textit{Keyframes.} For all other steps, the motion generator produces the intermediate states of the motion in the form of keyframes. Each keyframe specifies, for the participating robots, the torso position and orientation, and the world coordinates and contact label of each foot. Keyframes are generated only at the intermediate points of the motion where all four feet are in contact.

The code-based modules solve the joint angles of every keyframe with inverse kinematics, and check joint limits, foot support, and collisions.

\begin{figure}[t]
\centering
\includegraphics[width=\linewidth]{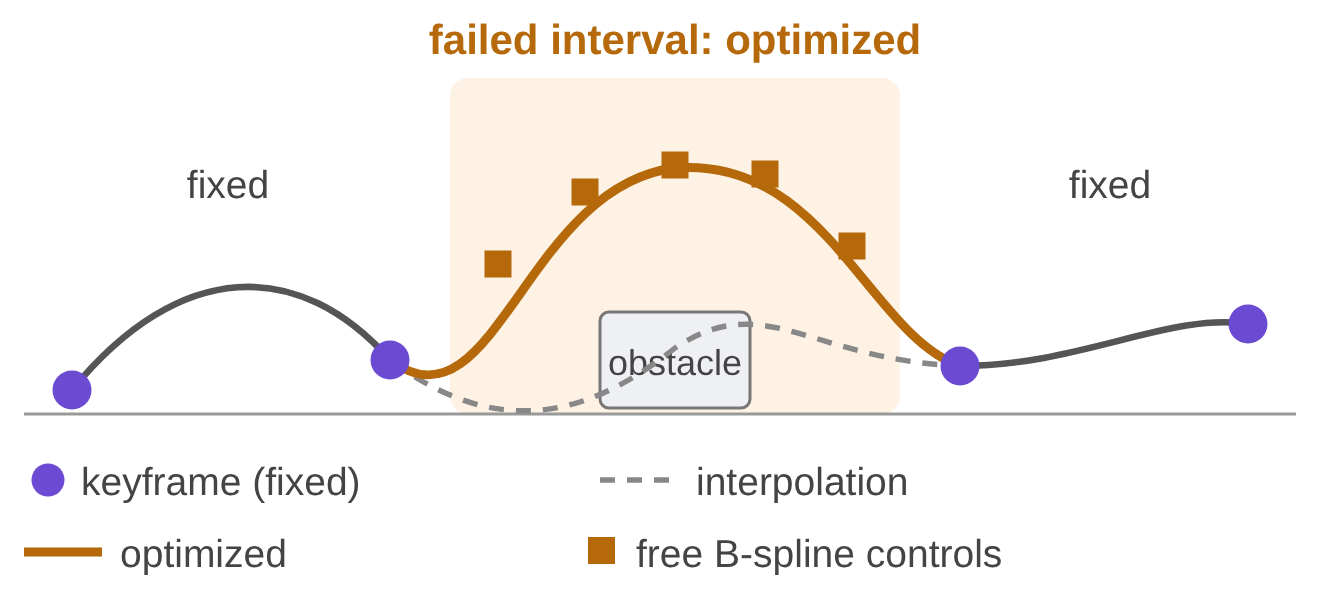}
\caption{Local kinematic optimization: only the failing interval between fixed keyframes is optimized.}
\label{fig:repair}
\end{figure}

\begin{figure*}[t]
\centering
\includegraphics[width=\linewidth]{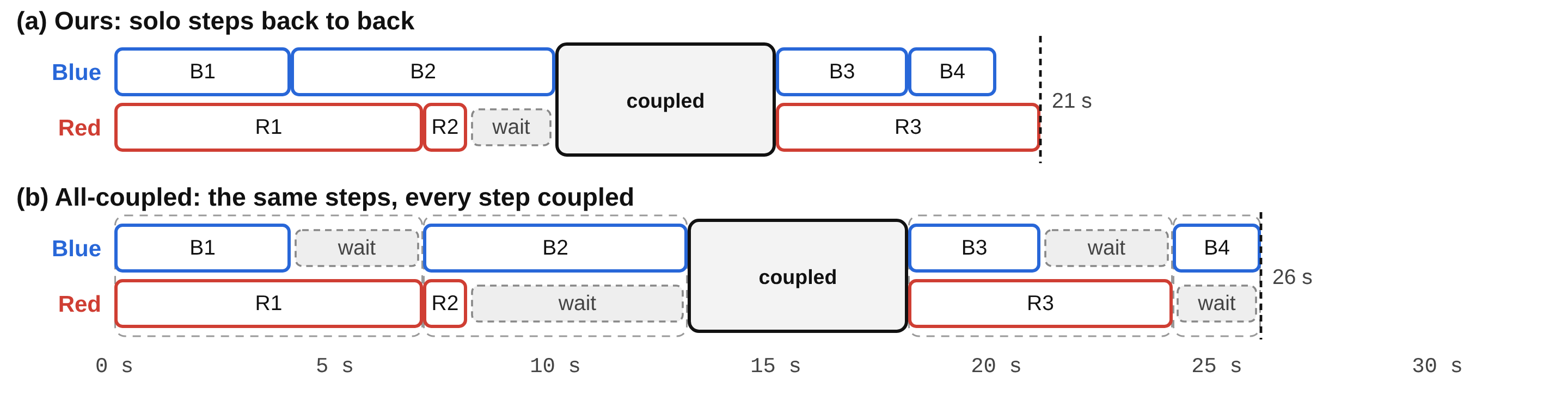}
\caption{Composition on a global timeline (schematic). The same steps are placed (a) as solo and coupled steps (Ours) and (b) as coupled steps only (All-coupled), in which the robot with the shorter motion waits at every step boundary.}
\label{fig:schedule}
\end{figure*}

\subsection{Kinematic Synthesis}
\label{sec:synthesis}

\textit{Interpolation.} The torso position and orientation are interpolated with cubic Hermite curves through the keyframes. Stance feet are kept fixed to their supports, and moving feet are interpolated to swing along half-elliptical paths. When a robot pushes a box, the box moves according to the motion of the link in contact with it.

\textit{Timing.} The time interval between consecutive keyframes is set adaptively from the body displacement and the swing distance of the feet within the interval. This keeps a natural motion speed throughout the trajectory.

\textit{Kinematic optimization.} Interpolation between valid keyframes can still violate constraints in between, for example when a swing foot grazes a step edge, an intermediate pose is unreachable, or a knee enters the partner's body. In that case, the code-based modules set an interval around the failed part and generate a path that avoids collisions and other violations through B-spline-based optimization (Fig.~\ref{fig:repair}).

\subsection{Validation, Correction, and Replanning}
\label{sec:correction}
The synthesized trajectory is checked every \SI{10}{ms} for joint limits, foot support, collisions, and support-edge margins; the dynamics are handled later by the TO stage (Sec.~\ref{sec:to}).

When a check fails, the failed items are returned to the motion generator together with renderings of the failed keyframes or trajectory segments. The motion generator then regenerates the keyframe sequence; this correction is repeated at most twice within the same session.

If a step still fails after two corrections, the plan itself may have specified a motion that cannot be realized, for example a single robot that is planned to climb an obstacle higher than it can reach on its own. In that case repeating motion generation does not help, so the failure and its diagnostics are returned to the task planner, which revises the plan, for instance by replacing the solo climb with a cooperative motion in which the partner provides support; this replanning is performed at most once.

\section{Trajectory Integration}
\label{sec:to}
\label{sec:composition}
The validated step motions are integrated into one trajectory for the whole task in two steps.

\textit{Composition.} Each step trajectory is generated separately, so the step trajectories are placed on one shared timeline to form the trajectory of both robots for the whole task (Fig.~\ref{fig:schedule}a). Each robot executes its own steps one after another without pause, and it waits only in two cases: when a precedence constraint requires a step of the other robot to finish first, and when its next step is a coupled step and the partner is not yet ready. While waiting, the robot stands still in its last pose.

\textit{Trajectory optimization.} The composed kinematic reference fixes the contact sequence and serves as the initial guess of a TO over both robots, formulated with Pinocchio~\cite{carpentier2019pinocchio} and CasADi~\cite{andersson2019casadi} and solved with IPOPT~\cite{wachter2006ipopt}. The TO finds a trajectory that stays close to the reference while satisfying the constraints of both robots and the pushed boxes. The constraints include the articulated rigid-body dynamics, contact forces within friction cones, the reciprocal forces between robots in contact and between robots and boxes, joint and torque limits, and collision avoidance. The boxes are modeled as passive bodies sliding along their rails. Because a long trajectory has too many variables to be solved at once, the course is divided into time windows that are solved in sequence, and consecutive windows are joined by matching their boundary conditions.

\section{Experimental Setup}
\label{sec:experiments}

\begin{table}[t]
\centering
\caption{Results of Our Framework (10 Trials per Course; Generation Time Excludes TO)}
\label{tab:results}
\footnotesize
\setlength{\tabcolsep}{4pt}
\begin{tabular}{lccccc}
\toprule
 & \multicolumn{2}{c}{Success} & Gen. time & Cost & Length \\
Course & Kin. & TO & [min] & [USD] & [s] \\
\midrule
1 & 10/10 & 10/10 & 15.3 & 1.68 & 52.9 \\
2 & 10/10 & 10/10 & 14.0 & 1.72 & 58.0 \\
3 & 10/10 & 10/10 & 13.5 & 2.46 & 70.1 \\
\bottomrule
\end{tabular}
\end{table}

\subsection{LLM Settings}
All LLM modules use Codex CLI (version 0.167.2)~\cite{codexcli} with \texttt{gpt-6.1-sol} at medium reasoning effort. Our programs call Codex in its non-interactive mode (\texttt{codex exec}), with approval requests disabled and file writes restricted to the module's workspace, and correction requests resume the same session. Within its workspace, a module can freely write and run code while solving its part, and it returns a JSON file in a prescribed format. Each call is additionally wrapped in a bubblewrap sandbox~\cite{bubblewrap} that hides everything outside the module's workspace, so a module cannot read the repository, the code that checks it, or the results of other steps and trials. Every trial starts fresh sessions; trials do not share responses or generated artifacts, and all inputs are frozen and verified by hashes.

\subsection{Experiments}

We answer five questions.
\begin{itemize}
\item \textit{Q1:} Does the proposed framework successfully perform cooperative TAMP?
\item \textit{Q2:} Does the proposed framework outperform a minimal single-call LLM baseline (Barebone)?
\item \textit{Q3:} Does the solo/coupled decomposition shorten execution by reducing unnecessary waiting (All-coupled)?
\item \textit{Q4:} Is keyframe-based motion generation more effective than having the LLM generate the full trajectory directly (Full-trajectory)?
\item \textit{Q5:} Does the kinematic optimization raise the success rate of kinematic synthesis for difficult motions?
\end{itemize}

\subsection{Metrics}
\emph{Kinematic success} means that the composed reference passes the dense validation of Sec.~\ref{sec:correction} and both robots reach their goals. \emph{TO convergence} means that, for a kinematically successful reference, the TO of Sec.~\ref{sec:to} converges to a trajectory that satisfies its constraints. Convergence suggests that the generated reference is dynamically plausible. The generation time is the wall-clock time from the start of a trial until a valid kinematic reference exists, excluding the TO. The steps are generated in parallel, so the motion generation stage counts as the duration of its longest step. Cost is computed from the token counts at OpenAI's published standard-tier API prices for \texttt{gpt-6.1-sol}: \$2.00, \$0.10, and \$10.00 per million uncached input, cached input, and output tokens~\cite{openaipricing}. Execution time is the total duration of the generated trajectory.

\section{Results}
\label{sec:results}

\begin{figure}[t]
\centering
\includegraphics[width=\linewidth]{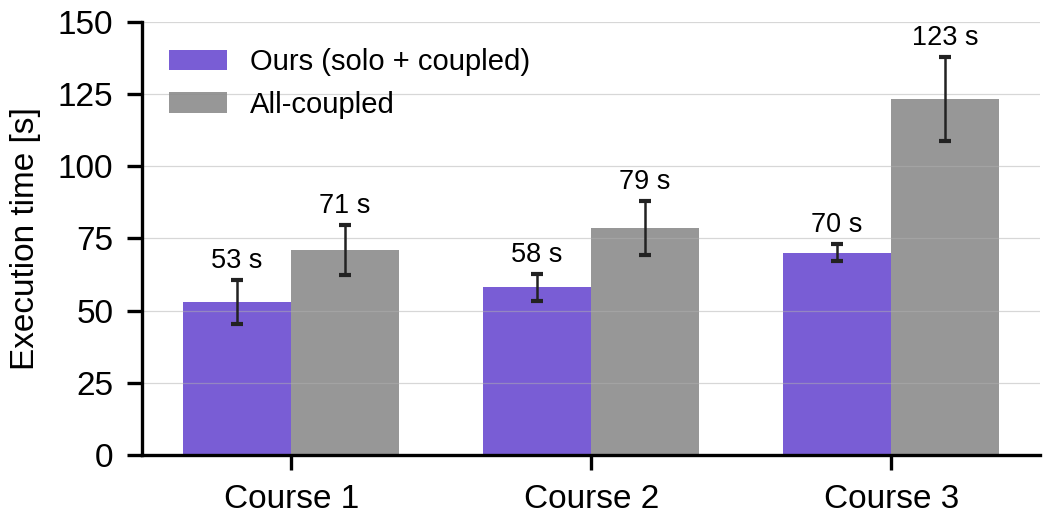}
\caption{Execution time with the solo/coupled decomposition (Ours) and with every step coupled (All-coupled), successful trials of 10 per course (mean $\pm$ s.d.).}
\label{fig:makespan}
\end{figure}

\subsection{Whole-Course Performance (Q1)}
We run the full pipeline ten times per course from scratch, with up to two corrections per step and at most one replanning. 
Table~\ref{tab:results} summarizes the ten runs per course. Our framework produced a kinematically valid reference in all 30 trials, and the TO converged in all of them. Generating the kinematic reference took 13.5--15.3~min per course on average. Because the steps of a plan are independent of each other, their motions are generated in parallel, which greatly reduces the generation time. The resulting motions last 53--70~s. The order and details of the motions differed slightly from trial to trial, but every trial produced a valid trajectory that brought both robots to their goals.

\subsection{Comparison with Barebone (Q2)}
To show that our framework is better than simply calling an LLM, we set up a minimal baseline, \emph{Barebone}. It receives the same inputs as the task planner (Sec.~\ref{sec:planning}), but with only a minimal prompt, it generates the full trajectory of the generalized coordinates of both robots and all boxes in a single LLM call. 
Table~\ref{tab:barebone} compares the success rates of our framework (Ours) and Barebone. Barebone produced kinematically valid trajectories in 6/10, 7/10, and 10/10 trials on the three courses. On Courses~1 and~2, however, most of these trajectories failed in TO (0 of 6 and 2 of 7 converged), because they contained physically implausible motions such as jumps. On Course~3, where the robots cooperate by pushing boxes without touching each other, the TO success rate was relatively high (8 of 10). Our framework converged in all trials, so the gap appears in motions where the robots physically support each other. This is also why we report kinematic success and TO convergence separately.

\begin{table}[t]
\centering
\caption{Success Rates of Ours and Barebone (10 Trials per Course; Barebone Without the Support-Edge Margin; TO Counted over Kinematically Valid Trajectories)}
\label{tab:barebone}
\footnotesize
\begin{tabular}{lcccc}
\toprule
 & \multicolumn{2}{c}{Ours} & \multicolumn{2}{c}{Barebone} \\
Course & Kin. & TO & Kin. & TO \\
\midrule
1 & 10/10 & 10/10 & 6/10 & 0/6 \\
2 & 10/10 & 10/10 & 7/10 & 2/7 \\
3 & 10/10 & 10/10 & 10/10 & 8/10 \\
\bottomrule
\end{tabular}
\end{table}

\subsection{Solo/Coupled Decomposition (Q3)}
To show the effect of planning solo and coupled steps separately and composing them afterwards, we set up an \emph{All-coupled} baseline that generates every step as a coupled step; the rest of the pipeline is the same. 

\begin{table}[t]
\centering
\caption{Success and Generation Time of Ours and All-Coupled (10 Trials per Course)}
\label{tab:allcoupled}
\footnotesize
\begin{tabular}{lcccc}
\toprule
 & \multicolumn{2}{c}{Kin. success} & \multicolumn{2}{c}{Gen. time [min]} \\
Course & Ours & All-coupled & Ours & All-coupled \\
\midrule
1 & 10/10 & 10/10 & 15.3 & 26.3 \\
2 & 10/10 & 10/10 & 14.0 & 18.0 \\
3 & 10/10 & 9/10 & 13.5 & 16.7 \\
\bottomrule
\end{tabular}
\end{table}

Fig.~\ref{fig:makespan} compares the execution times over ten trials per course. The All-coupled baseline solved 29 of the 30 trials. Compared with this baseline, our framework reduced execution time by approximately 25\%, 26\%, and 43\% on Courses~1--3, respectively (52.9, 58.0, and 70.1~s versus 71.0, 78.6, and 123.3~s). In the all-coupled plans the robot with the shorter part of a step waits for its partner at every step boundary (Fig.~\ref{fig:schedule}b), which lengthens the whole motion. The gap is largest on Course~3, where the two robots have long independent sequences of box pushes. Coupled steps are also harder to generate, since one LLM call must reason about both robots at once. Compared with the All-coupled baseline, our framework reduced generation time by approximately 42\%, 22\%, and 19\% on Courses~1--3, respectively (Table~\ref{tab:allcoupled}).

\begin{figure}[t]
\centering
\includegraphics[width=\linewidth]{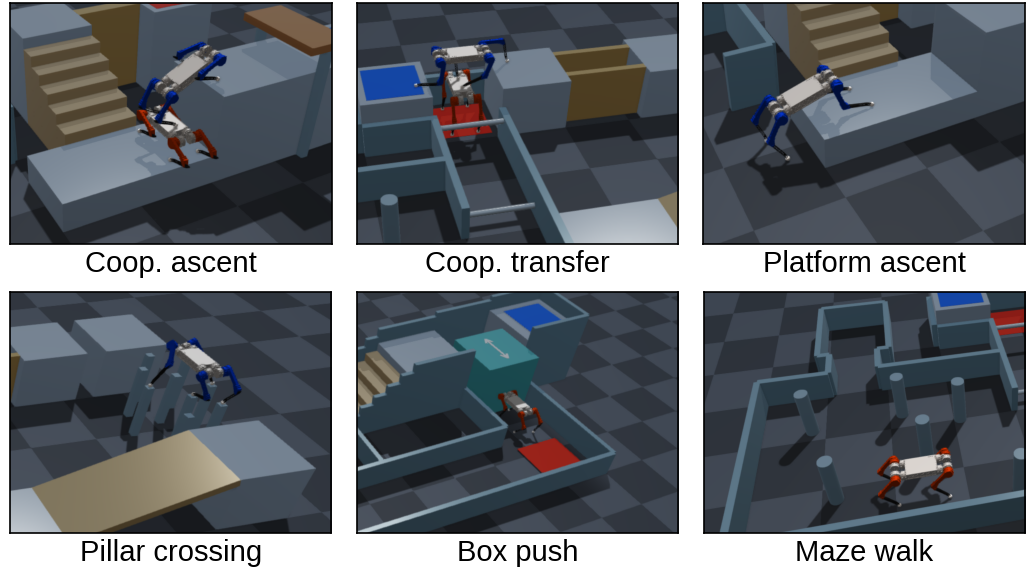}
\caption{The six steps of the step-level comparison. The first two are coupled steps, the others solo steps.}
\label{fig:steps}
\end{figure}

\begin{figure}[t]
\centering
\includegraphics[width=\linewidth]{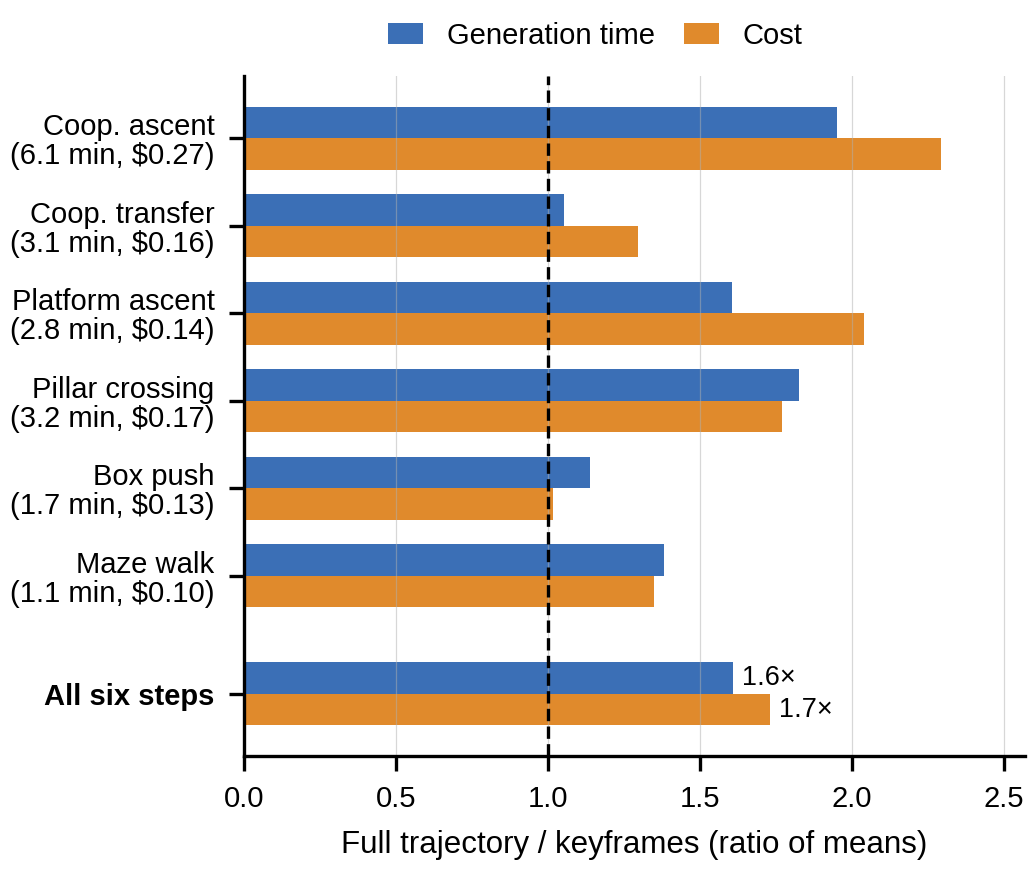}
\caption{Generation time and cost of full trajectories relative to keyframes (10 trials per step and method). Values in parentheses: mean generation time and cost of each step with keyframes. The dashed line indicates equal generation time or cost.}
\label{fig:kfvsfull}
\end{figure}

\begin{table}[t]
\centering
\caption{Kinematic Synthesis Without and With the Optimization (Same Keyframes, 10 Trials per Step)}
\label{tab:repair}
\footnotesize
\setlength{\tabcolsep}{4pt}
\begin{tabular}{llcc}
\toprule
Step & Characteristic & Without & With \\
\midrule
Coop. ascent & height, partner & 0/10 & 10/10 \\
Coop. transfer & height, partner & 2/10 & 10/10 \\
Platform ascent & height & 0/10 & 10/10 \\
Pillar crossing & flat & 8/10 & 10/10 \\
Box push & flat & 10/10 & 10/10 \\
Maze walk & flat, walking & 10/10 & 10/10 \\
\midrule
Total & & 30/60 & 60/60 \\
\bottomrule
\end{tabular}
\end{table}

\subsection{Keyframes vs.\ Full Trajectories (Q4)}
To show that the keyframe-based motion generation of our framework is efficient in time and cost, we set up a \emph{Full-trajectory} baseline. It uses the same inputs and validation code as our motion generator, but instead of outputting keyframes that are then synthesized, the LLM directly generates the dense full trajectory. For the comparison, we keep a successful plan fixed and repeat only the motion generation stage, ten times per method, on six challenging steps of that plan (Fig.~\ref{fig:steps}). 
On the six steps both representations eventually succeeded in 60/60 trials, so the difference lies in efficiency (Fig.~\ref{fig:kfvsfull}). Compared with the Full-trajectory baseline, keyframe-based generation reduced generation time by approximately 38\% and cost by approximately 41\%. The gap grows with the length of the contact-rich motion: on the coupled ascent, keyframes reduced generation time by approximately 50\% and cost by approximately 57\% compared with generating every \SI{10}{ms} sample directly, whereas the reductions were smallest for the box push and the cooperative transfer (approximately 0--23\%).

\subsection{Effect of the Kinematic Optimization (Q5)}
Q5 examines how the optimization in the kinematic synthesis affects the success rate of the interpolated reference. For the keyframes of the six steps of Q4, we run the kinematic synthesis twice, once without the optimization (interpolation only) and once with it, and compare how often the dense validation passes. Table~\ref{tab:repair} shows that the synthesis without the optimization succeeded in 30/60 cases and the synthesis with the optimization in 60/60. The effect concentrates on steps with height changes and robot--robot contact: on the cooperative and platform ascents, the synthesis without the optimization never succeeded, failing with IK failures and penetrations, whereas the maze walk and the box push were already valid. This shows that the kinematic optimization is highly effective for difficult motions with many contacts and a high risk of collision.

\section{Discussion and Limitations}
\label{sec:discussion}
The results show the benefit of letting the LLM modules and the code-based modules each do what they do best. The LLM modules make the high-level decisions, the plan and a few keyframes per step, while the code-based modules enforce the dense consistency through interpolation, optimization, and validation; Q4 and Q5 show the benefit of this split, and Barebone (Q2) shows what happens without it. Because the plan fixes the start and end states of every step, the steps are independent and their motions can be generated in parallel (Q1), and the solo/coupled decomposition additionally shortens the execution (Q3). Finally, kinematic validity does not imply dynamic feasibility: 13 of the 23 kinematically valid trajectories of Barebone did not converge in TO, whereas all references of our framework converged. Checking the dynamics with TO is therefore necessary.

The framework also has limitations. It currently handles only two robots, and cooperation among more robots would require coupled steps that involve more than two of them. Parallel generation greatly reduces the generation time, but obtaining a kinematic reference still takes about 15~min per course, mostly spent in LLM calls, which limits its use for fast replanning. Moreover, we have evaluated the framework only on three simulated courses, so its behavior in more diverse environments and on real hardware remains to be validated. 

\section{Conclusion}
\label{sec:conclusion}
We presented a framework for cooperative task and motion planning of two quadruped robots. LLM modules make the high-level decisions, namely the plan and the keyframes of each step, and code-based modules handle validation, interpolation, kinematic optimization, and composition. On three cooperative courses, our framework produced a kinematically valid trajectory in all 30 trials, taking about 15~min per course. A single LLM call that generates the whole trajectory rarely succeeded on the same courses. By reducing the time the robots wait for each other, the solo/coupled decomposition shortened execution time by approximately 25--43\% compared with the All-coupled baseline. Compared with generating dense trajectories directly, keyframes achieved the same success rate while reducing generation time by approximately 38\% and cost by approximately 41\%. The kinematic optimization raised the success of the kinematic synthesis from 30/60 to 60/60, mainly on motions with height changes and many contacts. Future work will extend the framework to more robots, reduce the generation time, and validate it in more diverse environments and on real hardware.

\bibliographystyle{IEEEtran}
\nocite{kiras2026}
\bibliography{refs}

\end{document}